\documentclass[conference]{IEEEtran}
\IEEEoverridecommandlockouts
\usepackage{cite}
\usepackage{amsmath,amssymb,amsfonts}
\usepackage{algorithmic}
\usepackage{graphicx}
\usepackage{textcomp}
\usepackage{xcolor}
\usepackage[colorlinks, urlcolor=cyan]{hyperref}
\usepackage[nolist,nohyperlinks]{acronym}
\def\BibTeX{{\rm B\kern-.05em{\sc i\kern-.025em b}\kern-.08em
    T\kern-.1667em\lower.7ex\hbox{E}\kern-.125emX}}

\usepackage{tikz}  
\usetikzlibrary{shapes.geometric, arrows}
\tikzstyle{node} = [circle, minimum width=1cm, text centered, draw=black, font=\small]
\tikzstyle{arrow} = [thick, ->, >=stealth, font=\small]

\usepackage{subcaption}
\newcommand{\modelfigheight}{3.7cm}
\newcommand{\demofigheight}{4.2cm}
\newcommand{\codefont}{\tt\small\verb}

\begin{acronym}
  \acro{ROV}{Remotely Operated Vehicle}
  \acro{TAN}{Terrain-aided navigation}
  \acro{GPU}{Graphics Processing Unit}
  \acro{ADCP}{Acoustic Doppler Current Profiler}
  \acro{UUV}{Unmanned Underwater Vehicle}
  \acro{DSL}{Deep Submergence Laboratory}
  \acro{DVL}{Doppler Velocity Logger}
  \acro{ROS}{Robot Operating System}
  \acro{WHOI}{Woods Hole Oceanographic Institution}
  \acro{USBL}{Ultra Short Base Line}
  \acro{LBL}{Long Base Line}
  \acro{SDF}{Simulation Description Format}
  \acro{INS}{Inertial Navigation System}
\end{acronym}

\IEEEoverridecommandlockouts
\IEEEpubid{\begin{minipage}[t]{\textwidth}\ \\[10pt]
        \centering\normalsize{Approved for public release. Distribution is unlimited.}
\end{minipage}} 

\begin{document}

\newcommand{\TODO}[1]{{\color{red}TODO: #1}}

%
\title{DAVE Aquatic Virtual Environment: Toward a General Underwater Robotics Simulator\\
}

\author{\IEEEauthorblockN{Mabel M. Zhang\textsuperscript{$\dagger$},
  Woen-Sug Choi\textsuperscript{$\ddagger$},
  Jessica Herman\textsuperscript{$\ddagger$},
  Duane Davis\textsuperscript{$\ddagger$},
  Carson Vogt\textsuperscript{$\ddagger$},
  Michael McCarrin\textsuperscript{$\ddagger$}, \\
  Yadunund Vijay\textsuperscript{$\dagger$},
  Dharini Dutia\textsuperscript{$\dagger$},
  William Lew\textsuperscript{$\dagger$},
  Steven Peters\textsuperscript{$\dagger$},
  Brian Bingham\textsuperscript{$\ddagger$}} \\
\IEEEauthorblockA{
  \textsuperscript{$\dagger$}\textit{Open Robotics}, Mountain View, California, USA \\
  {\tt\small\{mabel,
    yadunund,
    dharini,
    williamlew,
    scpeters\}@openrobotics.org} \\
  \textsuperscript{$\ddagger$}\textit{Naval Postgraduate School}, Monterey, California, USA \\
  {\tt\small\{woensug.choi.ks,
    jessica.herman,
    dtdavi1,
    crvogt,
    mrmccarr,
    bbingham\}@nps.edu}
}
}

\maketitle

\begin{abstract}
We present DAVE Aquatic Virtual Environment (DAVE)\footnote{DAVE is available at \url{https://github.com/Field-Robotics-Lab/dave}}, an open source simulation stack for underwater robots, sensors, and environments.
Conventional robotics simulators are not designed to address unique challenges that come with the marine environment, including but not limited to environment conditions that vary spatially and temporally, impaired or challenging perception, and the unavailability of data in a generally unexplored environment.
Given the variety of sensors and platforms, wheels are often reinvented for specific use cases that inevitably resist wider adoption.

Building on existing simulators, we provide a framework to help speed up the development and evaluation of algorithms that would otherwise require expensive and time-consuming operations at sea.
The framework includes basic building blocks (e.g., new vehicles, water-tracking Doppler Velocity Logger, physics-based multibeam sonar) as well as development tools (e.g., dynamic bathymetry spawning, ocean currents), which allows the user to focus on methodology rather than software infrastructure.
We demonstrate usage through example scenarios, bathymetric data import, user interfaces for data inspection and motion planning for manipulation, and visualizations.
\end{abstract}

\begin{IEEEkeywords}
underwater simulation, marine simulation, sensor simulation
\end{IEEEkeywords}

\section{Introduction}\label{sec:intro}

Simulation is a fundamental capability for development and evaluation of robotic applications. By providing a parameterized approximation of complex scenarios, simulators enable rapid testing of new robotic solutions and at low cost in repeatable environmental conditions while reducing the time, cost, and risk of physical deployment.  Despite the coarse environmental abstractions of simulation, system-level testing in simulation enables developers to identify problems without solely relying on costly physical testing~\cite{timperley18crashing_url}.

\begin{figure}[ht]
    \centering
    \includegraphics[width=0.48\textwidth]{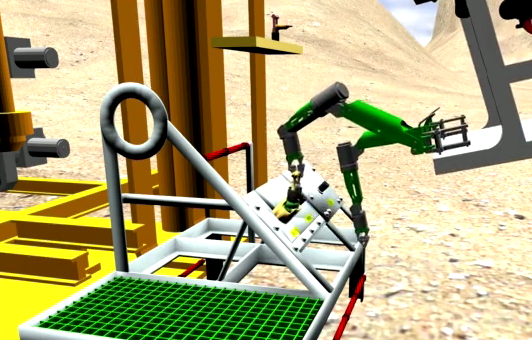}
    \caption{Bimanual plug insertion in the integrated scenario.}
    \label{f:cover}
    \vspace{-5mm}
\end{figure}

Robotic solutions are often envisioned as best suited for applications that have the three Ds: dirty, dangerous, and dull~\cite{takayama2009beyond_url}.  Analogously, simulation has particular value when scenarios have three Rs: remote, risky, and recalcitrant.  Because physical testing in remote operational areas such as the deep ocean, underground \cite{patchou2020hardware_url, rouvcek2019darpa_url}, and outer space \cite{hambuchen2017nasa_url, allan2019planetary_url, tanaka2017toward_url} is particularly costly, simulation is especially valuable as a complement to field deployments with limited access.  Similarly, simulation testing at multiple levels of fidelity is necessary for high-risk deployments where the financial risk of field campaigns is high, there is danger to human operators, or there is potential loss of robotic platforms~\cite{fountain2010abe_url}.

Finally, for recalcitrant environments that cannot be easily reproduced in field venues, simulation complements physical testing.  For example, the operating requirements for an ocean surface robot may include a specific envelope of environmental conditions (water current, waves, and wind).  Physical test and evaluation of the system throughout such an envelope is intractable.  Moreover, deep-ocean robotic scenarios are remote, often requiring weeks or months at sea aboard oceanographic support vessels and large logistical requirements for deployment and recovery.  Such operations involve significant risk due to limited weather windows where undiscovered software bugs have the potential to cause the loss of multi-million dollar robotic vehicles.  Furthermore, the environments are recalcitrant in the sense that the software must be designed for a wide range of oceanographic conditions, but operators do not have the ability to control the environmental conditions in which they are deployed.

This paper describes the open-source DAVE simulation environment which extends existing underwater robotic simulation capabilities to support the rapid testing and evaluation of solutions for deep-sea autonomous robotic intervention activities.  The virtual environment consists of models of common operating scenarios and objects, capabilities to emulate environmental influences on vehicle physics and sensors, and tools for generating authentic, sensor-based scenarios.  The simulation environment approximates some of the distinct challenges of autonomous underwater mobile manipulation, enabling demonstration of the following critical aspects of potential solutions:
\begin{itemize}
\item Whole-body motion planning and position retention during operation in work area
\item Perception via multimodal sensor fusion for manipulation
\item Coordinated manipulation from hovering vehicle
\item Grasp planning, including risk and success assessment
\item Detecting completion and re-planning of manipulation
\end{itemize}

\section{Related work}\label{sec:rel_work}



Underwater simulation is not new.
Despite efforts in the past couple of decades \cite{mcmillan1996, choi2000, song2003, matsebe2008, sehgal2010, benjamin2018_moos_sim}, underwater simulators come and go, with no consensus on a lasting solution.
While their land counterpart enjoys an array of stable fully featured simulators, underwater robots are in a unique surrounding where even the environment poses a number of simulation overheads. 
As traditional simulators are not designed to handle the marine environment, it is customary for water features to be created by the user as add-ons.
Moreover, since neither hardware nor simulators have been standardized, the wheel is frequently reinvented to create simulators specific to each study.

Similar to previous efforts, we added marine features on top of an existing simulator designed for land robots, for its relatively long and stable existence.
Gazebo~\cite{koenig2004gazebo} is a comprehensive simulator considered the standard for various robotics applications and well-integrated with \ac{ROS}.
Its built-in capability for physics to be run at custom time steps makes faster-than-real-time execution easy.
DAVE builds on top of UUV Simulator~\cite{manhaes2016uuv}, a recent software stack that became a standard for underwater simulations, and \ac{DSL} {\codefont|ds_sim|}~\cite{vaughn_dssim}, by adding environmental and sensory features that are common building blocks for underwater, as well as tools and scenarios.

Prior to UUV Simulator's adoption, UWSim \cite{prats2012_uwsim} has been used for various applications \cite{kermorgant2014, fernandez2015, gwon2017}. However, it has been noted that while it is advantageous in visualization and sensors, it is lacking in dynamics simulation \cite{cieslak2019}.
UW MORSE \cite{henriksen2016_uwmorse} is built upon MORSE, a generic, though dated, simulator 
featuring \ac{USBL} and \ac{LBL} positioning systems, acoustic range sensors and a pressure depth sensor, among other capabilities.
Unfortunately, MORSE is no longer developed.
Around the same time, UUV Simulator \cite{manhaes2016uuv} became the de facto choice, with more sensors, scenarios, and general capabilities. DAVE includes everything from UUV Simulator and more.


Besides Gazebo, game engines known for photo-realistic rendering have also been chosen as the bases for underwater simulation.
Unreal Engine \cite{unreal} has been used in vision-based applications \cite{manderson2018_unreal, jamieson2021_unreal, potokar2022_unreal}.
Unity \cite{unity} has been used in simulators such as URSim \cite{katara2019_unity}.
These graphics- and game-driven simulators trade off physical fidelity with advanced rendering, a necessary balance for all simulators with finite computing resources.
Gazebo, on the other hand, prioritizes physics over rendering, offering higher physical confidence.

Deep dives into customized rendering independent of existing renderers also exist.
Stonefish \cite{cieslak2019} developed its own simulation framework, using OpenGL directly to achieve realistic rendering for surface and underwater.
Suzuki and Kawabata \cite{suzuki2020} developed an underwater simulator based on a generic platform called Choreonoid, simulating underwater cameras by adding noise, discoloration, and distortion.
While DAVE does not emphasize rendering, we note that other add-on options for photo-realistic rendering exist for Gazebo.
For example, Song et al. \cite{song2020} implemented deep sea imaging with attenuation, reflection, backscatter, and artificial illumination, and integrated with UUV Simulator in Gazebo.


Typically, underwater simulators provide hydrodynamics, buoyancy, sensors, and actuators, which are necessary but insufficient for a fully featured environment.
For a well-rounded underwater simulation, it is imperative to operate in large worlds on the scale of kilometers, whereas typical robotics applications are on the scale of meters.
Building on top of \cite{vaughn_dssim}, DAVE provides capabilities to load large elevation maps and to dynamically spawn tiles only when necessary, conserving computational resources.

Additionally, to simulate realistic data for sensors and algorithms, DAVE demonstrates importing real-world high-resolution bathymetry and ocean current data from free online databases, which are read into simulated sensors like the \ac{DVL}.
In addition to environments, various demonstration examples are provided, including vehicle models, object models, and tools to assist the user in setting up complex multi-purpose scenarios.


\section{Environment and tools}


The building blocks of underwater simulation start at the environment level. What for land is empty space is for underwater characterized by fluid, seabed, and their various natural features.
Moving beyond bathymetry heightmaps that define the seabed, we integrated a method for dynamic allocation of seabed tiles for scalability and performance.
Moving beyond constant currents, we offer ways to define 3D stratified ocean currents and tidal oscillations, both of which affect vehicle dynamics and sensing.
In these environments hover vehicles to perceive and interact with the world, and sit common objects subjected to environmental degradation.

\begin{figure*}[thbp]
    \begin{center}
        \begin{subfigure}{.36\textwidth}
                \includegraphics[height=\modelfigheight]{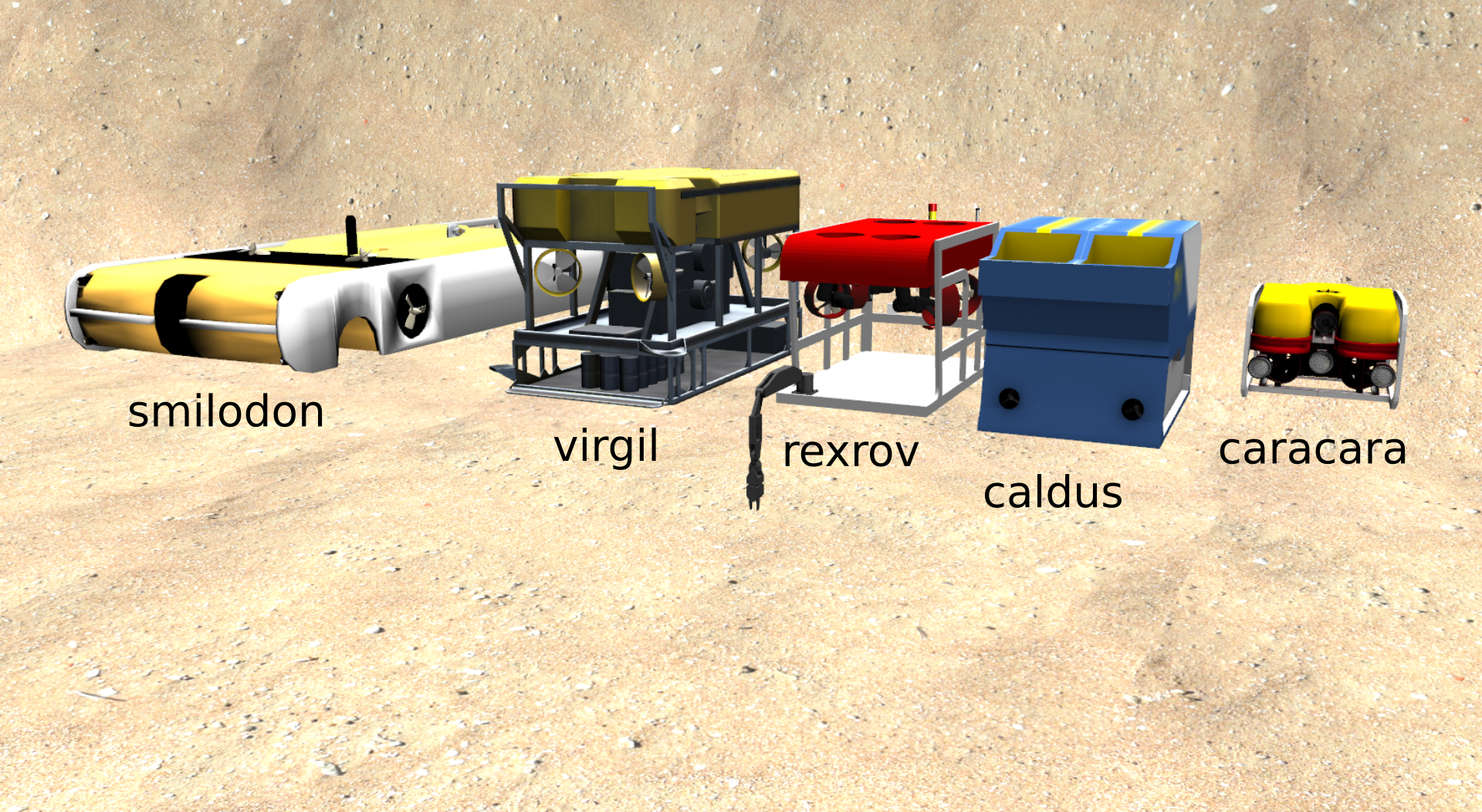}
            \caption{}
            \label{f:uuv_collection}
        \end{subfigure} \hfill
        \begin{subfigure}{.32\textwidth}
            \includegraphics[height=\modelfigheight]{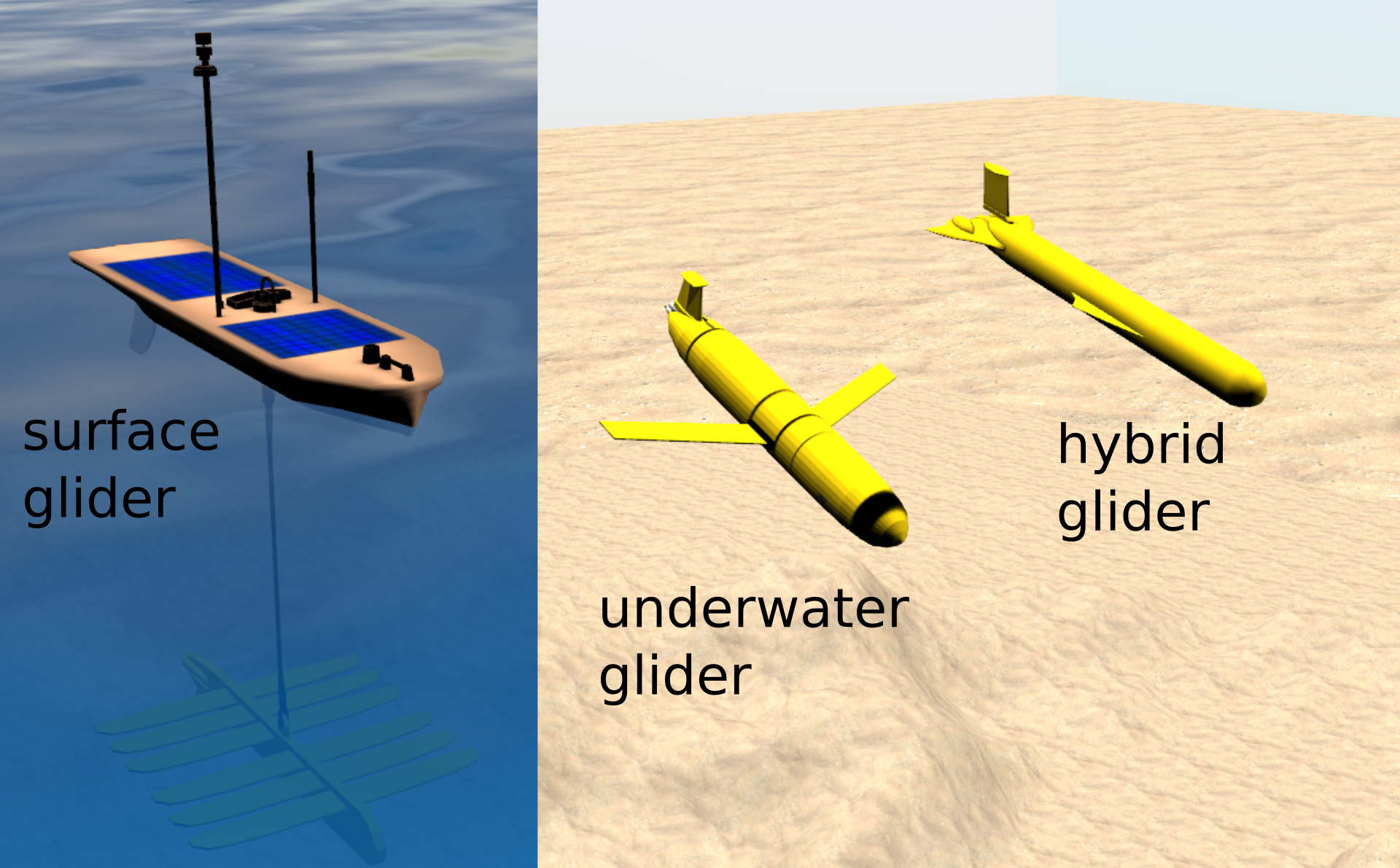}
            \caption{}
            \label{f:glider_collection}
        \end{subfigure} \hfill
        \begin{subfigure}{.30\textwidth}
            \includegraphics[height=\modelfigheight]{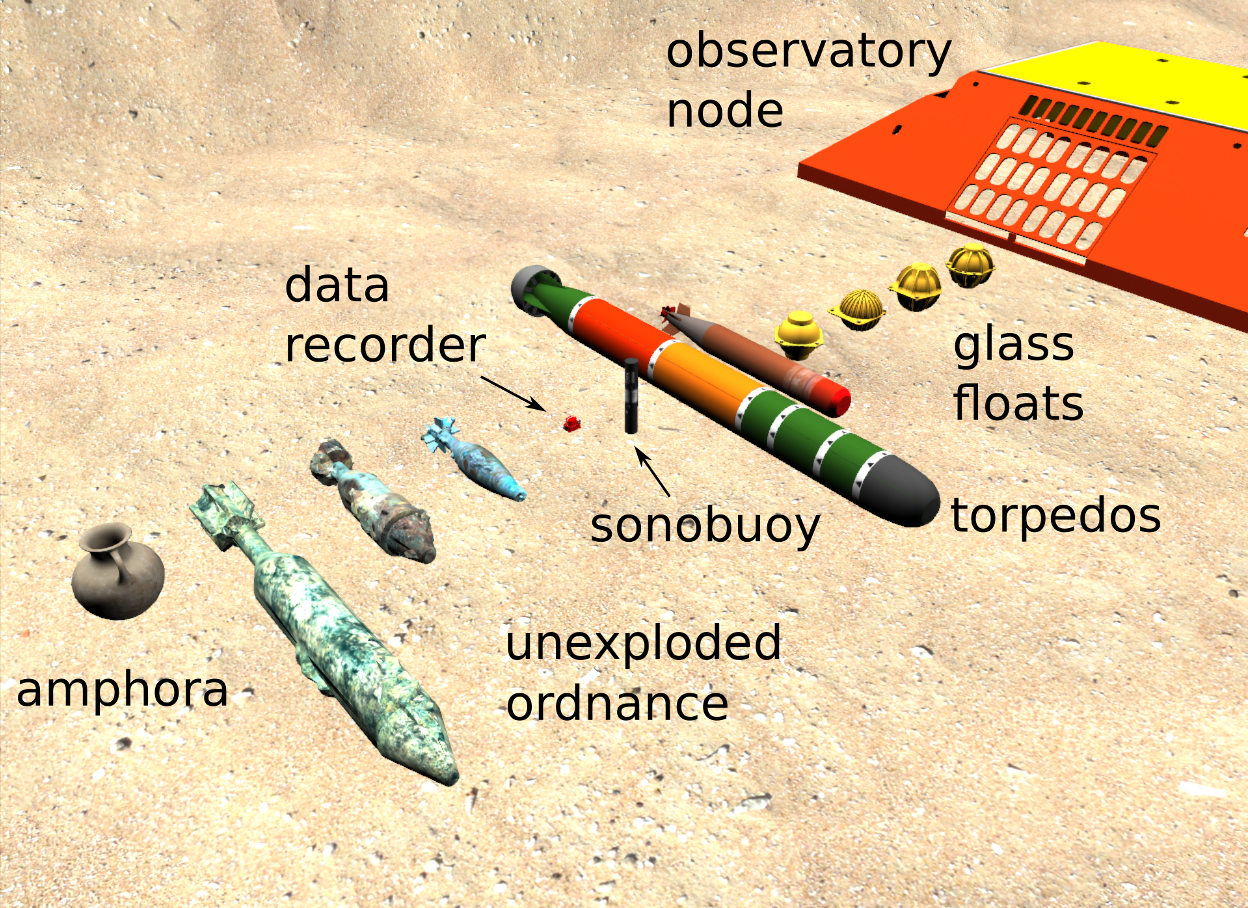}
            \caption{}
            \label{f:object_collection}
        \end{subfigure} \hfill%
    \end{center}
    \vspace{-5mm}
    \caption{(a) Vehicle models. (b) Glider models. (c) Object models.}
    \vspace{-5mm}
\end{figure*}

\subsection{Dynamic bathymetry spawning}\label{subsec:dynamic_bathy}


Bathymetry, the depth of the seafloor, is an essential component underwater. Operation areas are often on the scale of tens of kilometers, with resolution as fine as a few centimeters. This poses a challenge to memory consumption.
In DAVE, grid tiles produced from large, high-resolution bathymetry heightmap data are dynamically spawned and unloaded. This allows for efficient memory use and scales to very large worlds. A script is provided to generate these tiles, OBJ mesh files with colorization corresponding to the relative depth, from data.

Dynamic loading is implemented as a Gazebo world plugin, which inspects the geodetic coordinates of each vehicle and injects the tile needed on the fly. Regions are overlapped for continuity of sensor data readings. The geodetic coordinates, 
defined in the World Geodetic System (WGS 84; EPSG 4326), are calculated using the GDAL library~\cite{gdal}, while the Cartesian coordinates are described with Pseudo-Mercator (EPSG 3857), to support any bathymetry format and location.

\subsection{Ocean currents}\label{subsec:current}


Ocean currents are critical environmental characteristics for autonomous marine path planning and manipulation. Currents are spatially variant (stratified) in terms of water column and temporally variant due to tides. DAVE provides the framework to define stratified ocean currents by directions and depths, either constant or periodic in time. The tide cycle is described by a complete data set or harmonic constituents, calculated using the world clock. 

The currents are interpolated from the user-defined database and calculated using the first-order Gauss-Markov process model according to each vehicle's depth~\cite{fossen2011handbook}, $\dot{V}_{currents}+\mu V_{currents} = \omega$. Here, ${V}_{currents}$ is the velocity of the ocean current, $\omega$ is Gaussian noise, and $\mu\geq0$ (typically zero for white noise) is a constant.

The global ocean currents database, which can be modified on the fly, is published as \ac{ROS} and Gazebo messages from a world plugin. Its effect on each vehicle is calculated in a model plugin, for multi-vehicle support and water tracking in the \ac{DVL} (Section \ref{subsec:dvl}).

\subsection{Models}\label{subsec:models}


DAVE extends UUV Simulator's library of vehicle, manipulator, and object models for composing evaluation scenarios.

\subsubsection{Vehicle Models}
Following the usage patterns from UUV Simulator~\cite{manhaes2016uuv}, we added a number of surrogate robotic models representative of general classes of underwater robotic platforms (Fig.~\ref{f:uuv_collection}).  The RexROV test vehicle from UUV Simulator is shown in the middle for scale.  The collection of vehicles represent typical size and configurations of inspection and intervention class platforms.  Each platform includes the configuration of the thruster propulsion, hydrodynamic, and hydrostatic plugins described in~\cite{manhaes2016uuv}.


We also introduce surrogate platforms representative of three types of ocean gliders: surface gliders~\cite{hine09waveglider_url}, underwater buoyancy driven gliders~\cite{webb2001slocum_url} and underwater hybrid gliders~\cite{claus2010experimental_url} (Fig.~\ref{f:glider_collection}).  The motion model for these gliders can achieve faster-than-real-time simulation, which is essential for typical glider mission profiles.


\subsubsection{Object Models}

DAVE includes a library of representative items commonly associated with deep-sea detection, identification, and manipulation tasks.  A subset (Fig.~\ref{f:object_collection}) of the models are available on Gazebo Fuel\footnote{Gazebo Fuel online 3D models database: \url{https://app.gazebosim.org}}, a free online database of 3D models ready for plug and play in Gazebo.  Each model consists of a visual geometry mesh for rendering, a texture image overlaid on the mesh, a collision mesh used by the physics engine for interactions like grasping and manipulation, and inertial properties of each element for dynamics.



\subsection{Object distortion}\label{subsec:distortion}


Among other naturally occurring phenomena unique to the marine environment, physical degradation, including visual, geometric, and surface properties, is observed on objects submerged for prolonged periods \cite{biofouling2006}.
Algorithms need to be evaluated for robustness to such changes.

\subsubsection{Geometric distortion}

\begin{figure}[htbp]
    \centering
    \includegraphics[width=0.11\textwidth]{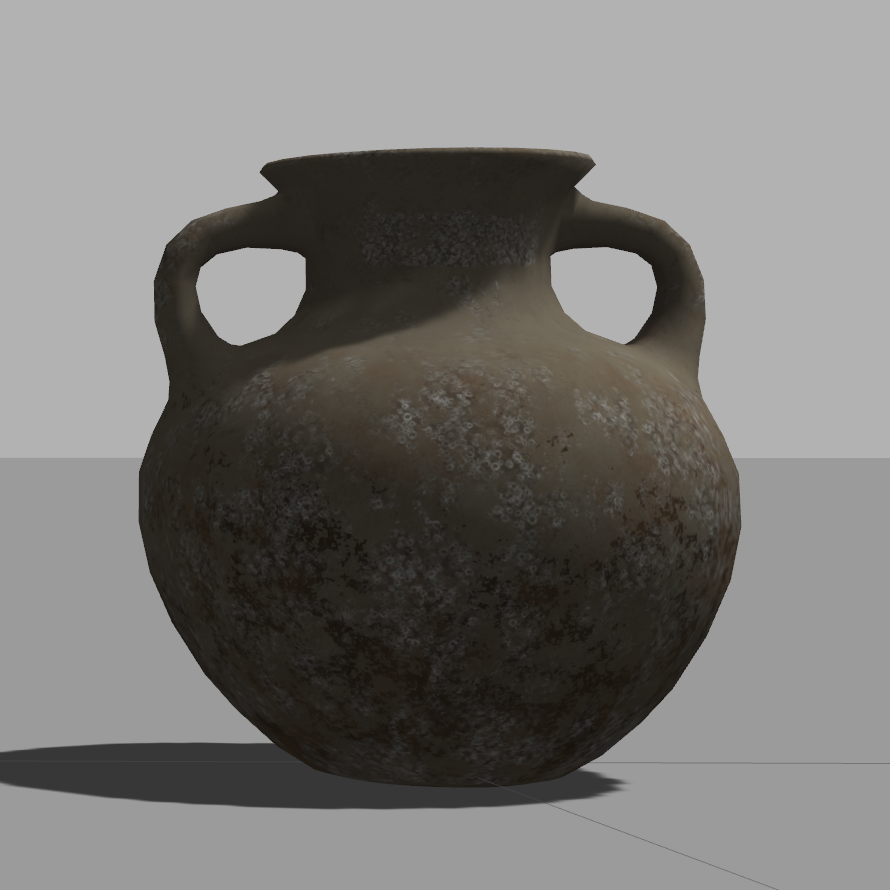}
    \includegraphics[width=0.11\textwidth]{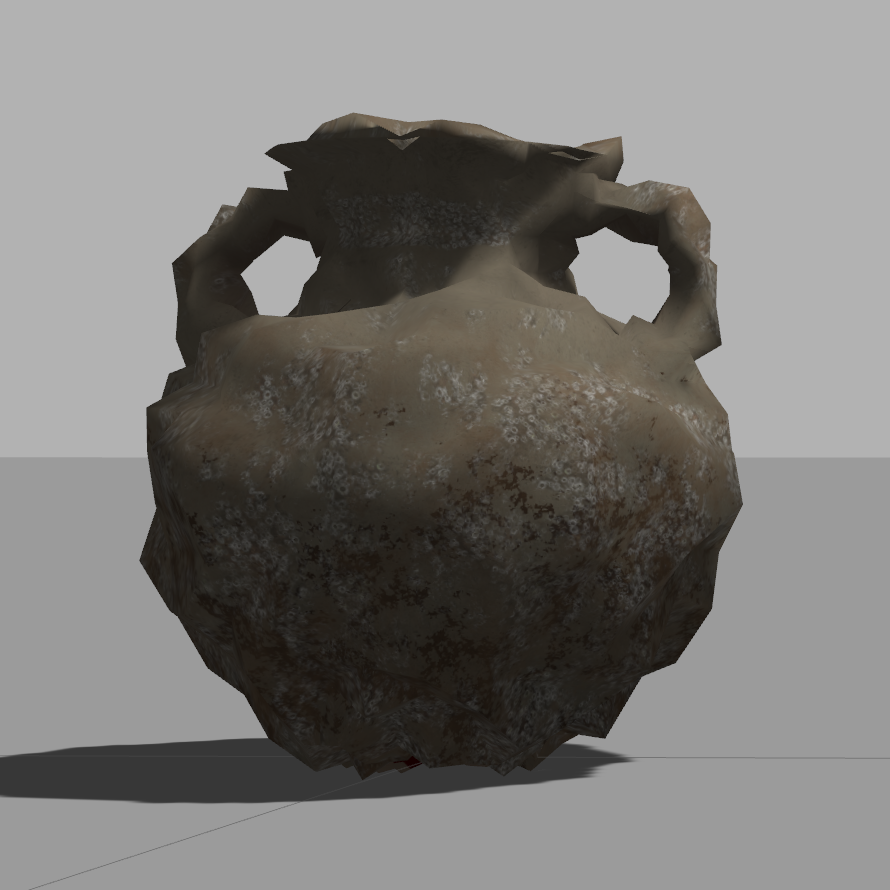}
    \includegraphics[width=0.11\textwidth]{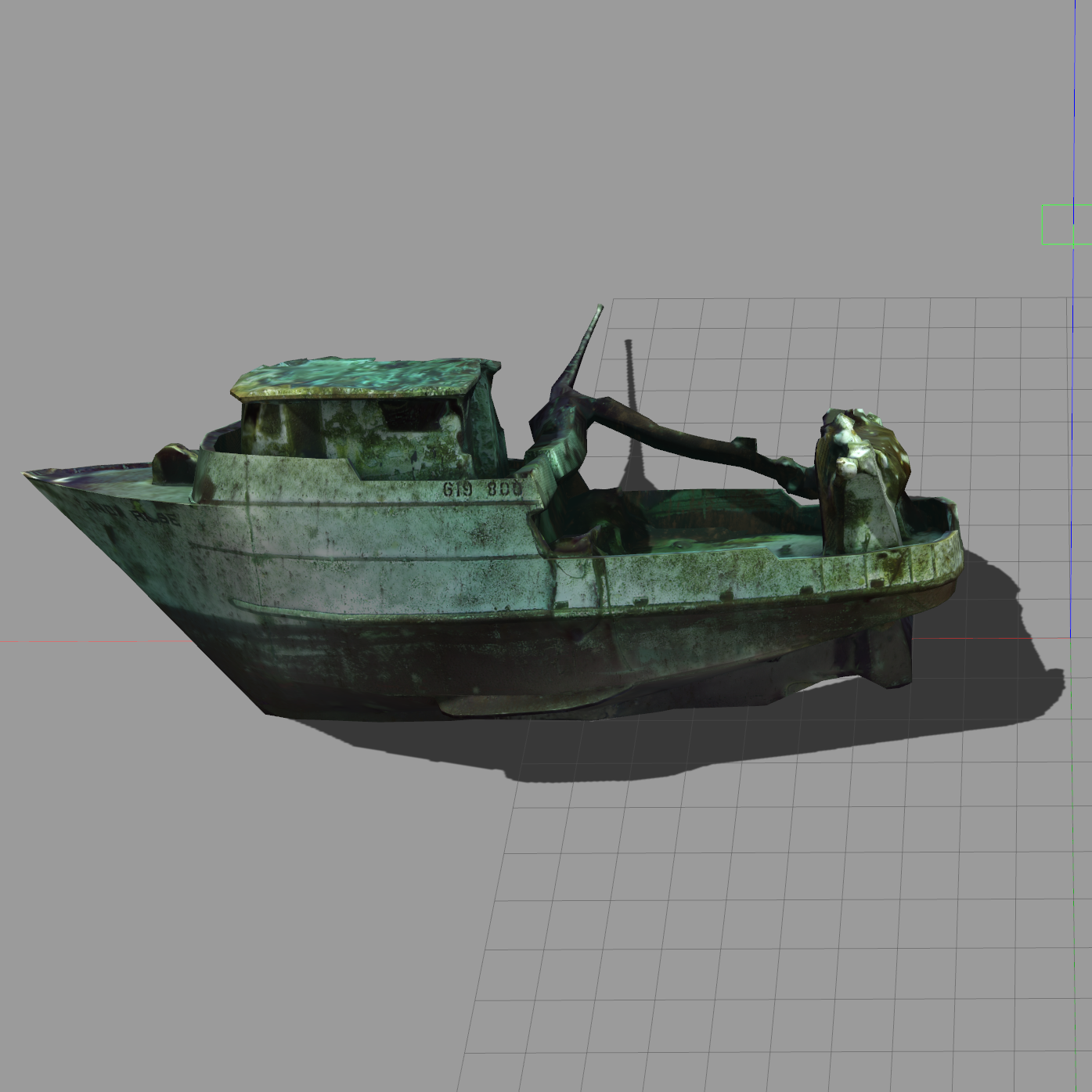}
    \includegraphics[width=0.11\textwidth]{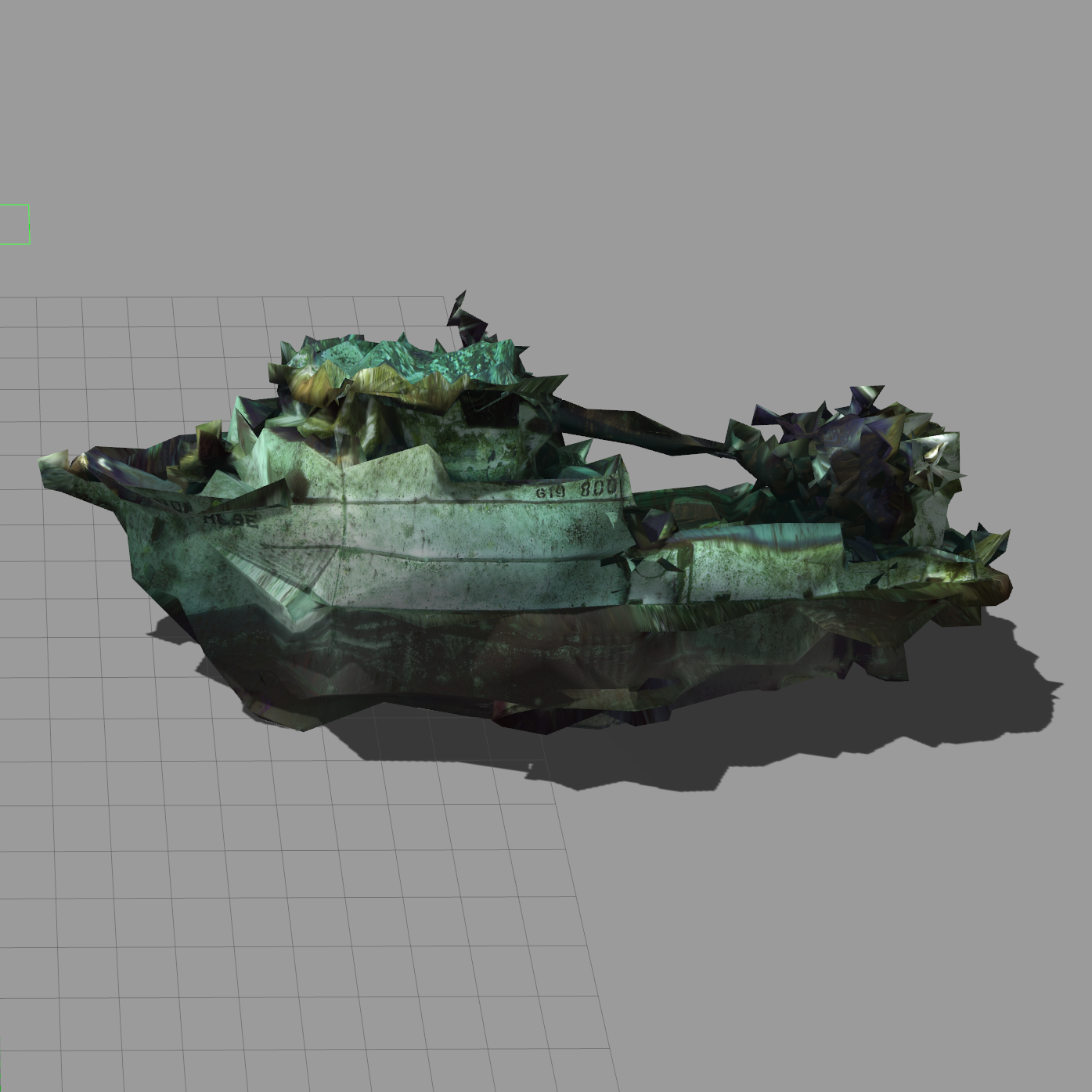} \\
    \includegraphics[width=0.11\textwidth]{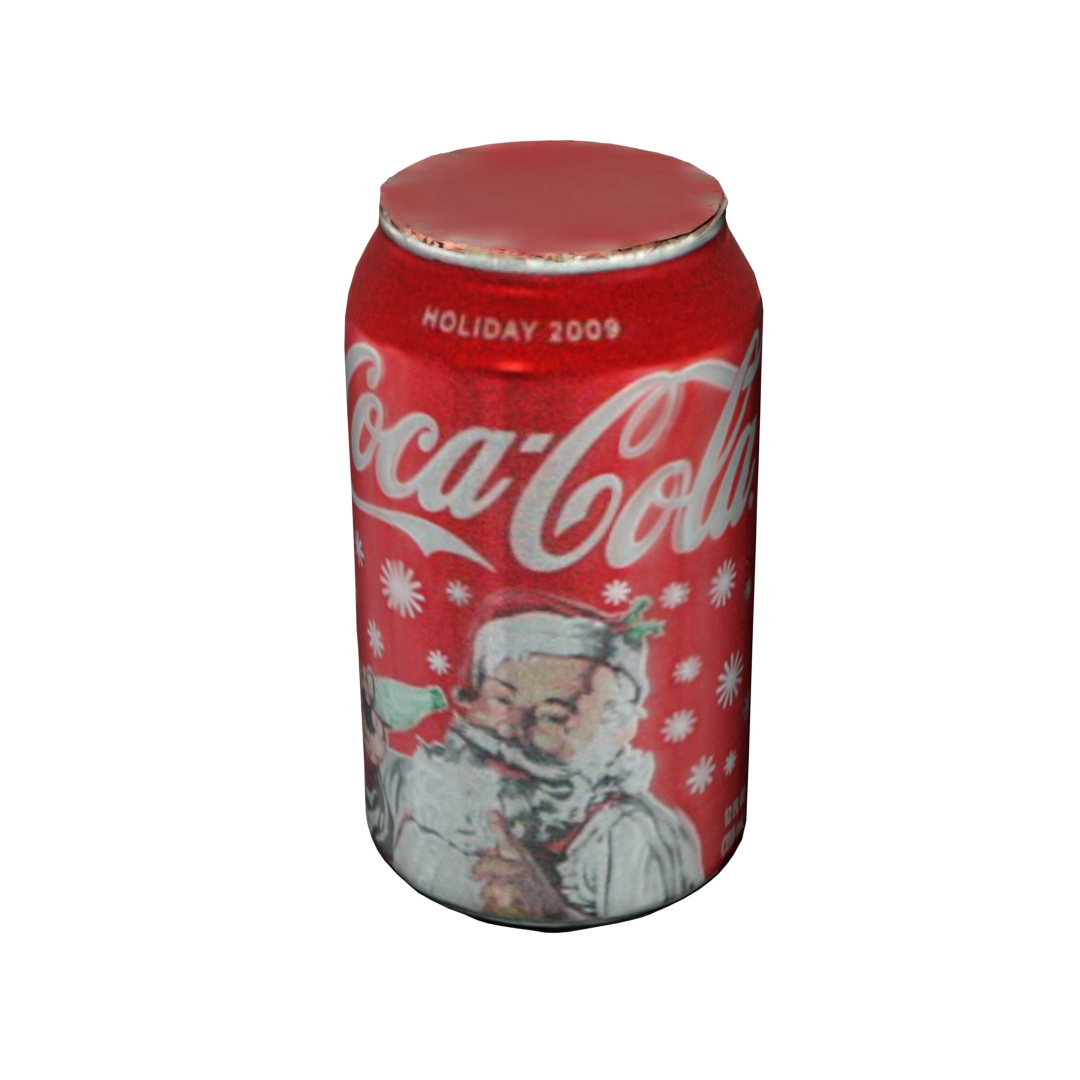}
    \includegraphics[width=0.11\textwidth]{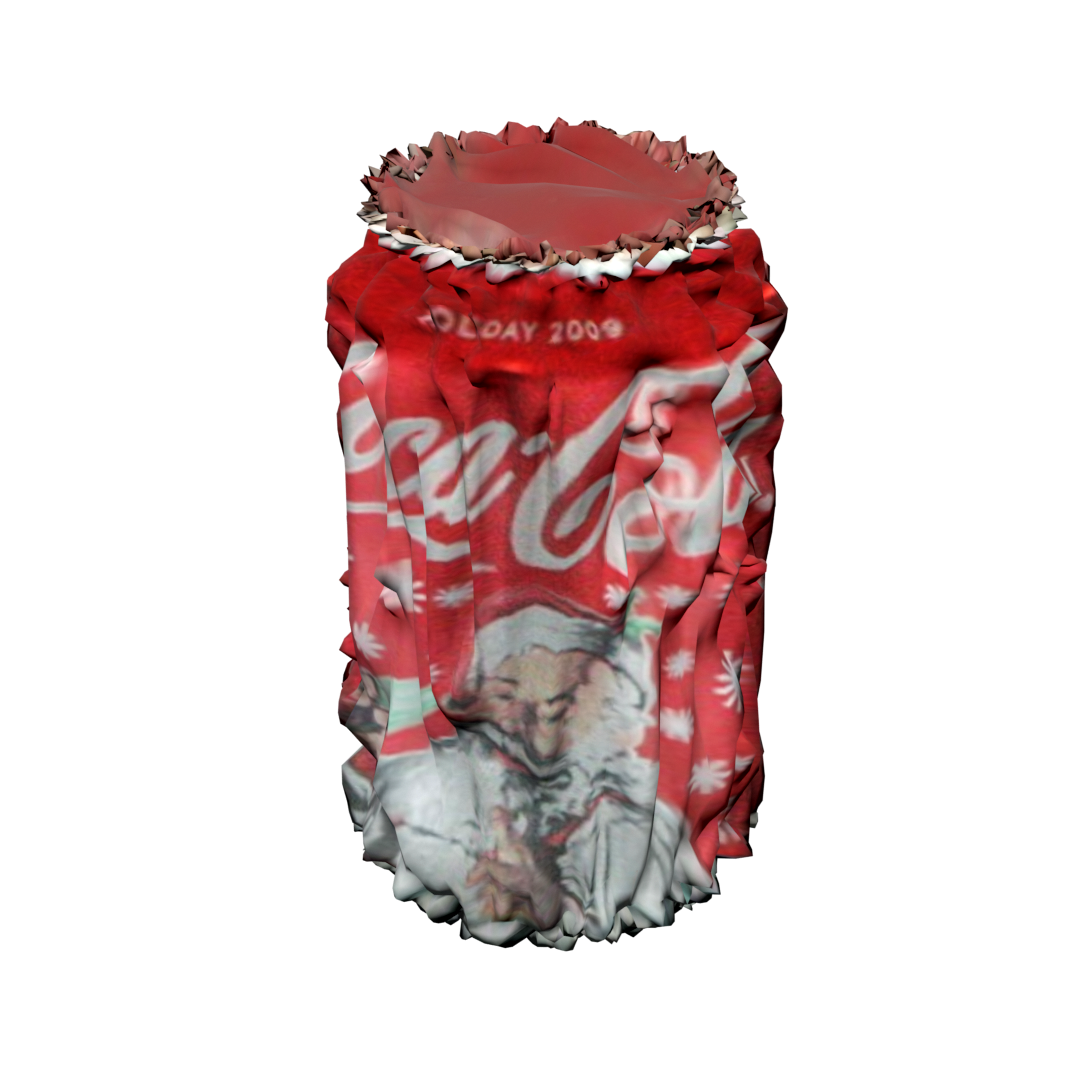}
    \includegraphics[width=0.11\textwidth]{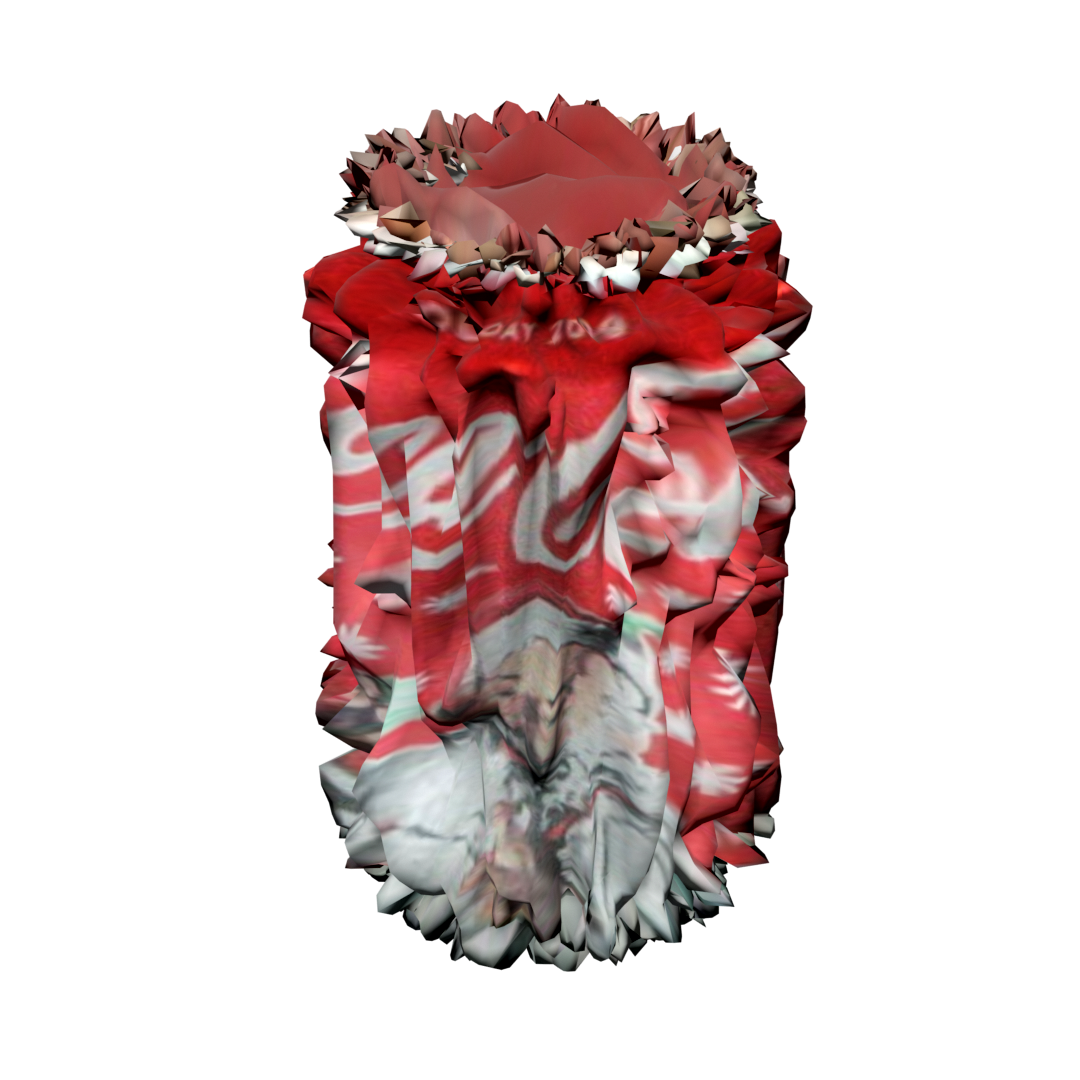}
    \includegraphics[width=0.11\textwidth]{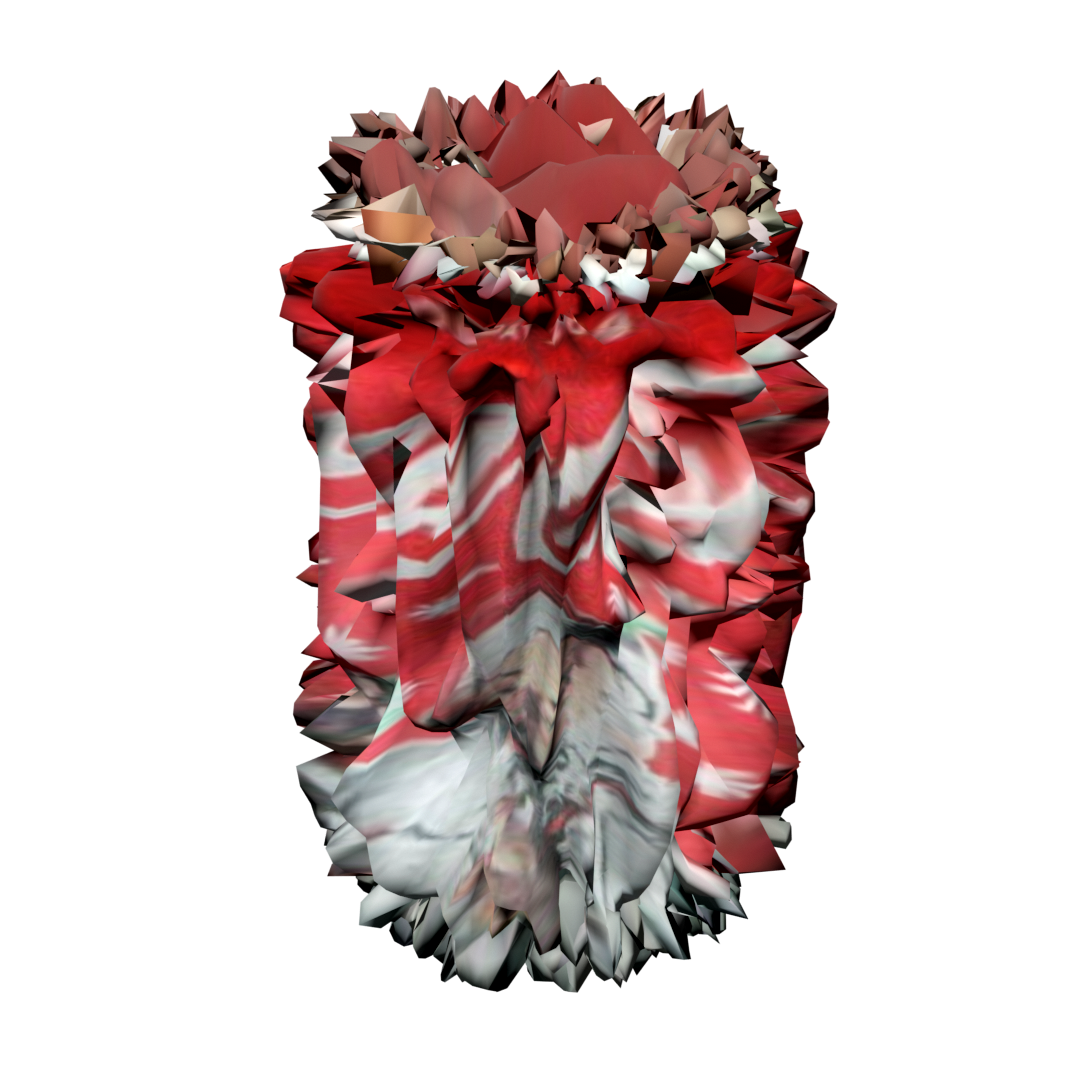}
    \caption{Example objects distorted to increasing extents.}
    \label{f:degradation}
\end{figure}

To speed up repetitive manual mesh distortion, programmatic mesh modifications can be used to facilitate quick and reproducible permutations.
One way is through Blender~\cite{blender}, a professional 3D modeling tool that exposes a full feature suite to its Python API.
We provide a sample Blender Python script as a starting point, demonstrating two types of distortions:
Subdivision Modifier splits edges and faces without changing physical appearance;
Randomized Vertices displace existing vertices by amounts bound by custom parameters, creating a deformed look.
The distortion extent is controlled by a parameter $\in [0, 1]$.
Fig.~\ref{f:degradation} shows objects distorted to various extents.

\subsubsection{Surface properties}

Friction coefficients, slip, and other parameters that affect physical interactions undergo changes during natural degradation.
In Gazebo, these parameters are specified in \ac{SDF}, an XML format.
Beyond those in the standard \ac{SDF} specification\footnote{SDF specification: \url{http://sdformat.org/}}, custom XML properties can be added.
We provide an \ac{SDF} generation script in Embedded RuBy (ERB) to demonstrate both standard and custom physical properties.
The aforementioned distortion extent parameter may be used here as an input.
We leave the calculation and tuning of the mathematical parameters to the user, as materials and surface properties vary greatly.
To make use of the custom properties in simulation, an example Gazebo plugin demonstrates parsing the custom XML at run-time.

\section{Sensors}\label{sec:sensors}

Underwater robots require different sensing modalities from land.
Vision underwater is severely impaired by scattering, refraction, and reflection, properties of water unfamiliar to land algorithms. As a result, the building blocks of underwater sensing consist of acoustics and ray casting, which penetrate much larger distances and are more efficient than optics.
Localization also requires unique sensing capabilities.
We extended existing and developed new simulations of common sensors.

\subsection{Doppler velocity logger (DVL)}\label{subsec:dvl}


\acp{DVL} are among the most common sensors for underwater navigation. The DAVE \ac{DVL} is based on the {\codefont|ds_sim|} Gazebo sensor and model plugin~\cite{vaughn_dssim}.  Characteristics like minimum and maximum range, beam orientations, and noise parameters are specified in the sensor and plugin \ac{SDF} elements.

The {\codefont|ds_sim|} sensor and plugin provided a bottom-track velocity in the sensor frame that is calculated from the relative velocities of the sensor and objects contacted by each \ac{DVL} beam. The sensor utilizes the Gazebo physics library to calculate the intersection between a Gazebo RayShape associated with each beam and the world model with which it collides. The relative velocities are converted to noise-free beam-specific scalar velocities. A noise-free sensor velocity solution is computed from the beam velocities and published as a Gazebo message that also includes beam velocities, ranges, and orientations.

To this \ac{DVL} solution, Gaussian noise is then added to each beam velocity, and the solution is recomputed in the sensor plugin.  The noisy velocity solution, sensor altitude, individual beam velocities, ranges, and orientations are published as a \ac{ROS} message. The sensor and plugin compute a bottom-tracking solution if at least three (of four) beams are in contact with the bottom.

DAVE extends the {\codefont|ds_sim|} \ac{DVL} by adding water tracking and current profiling capabilities using values published by the ocean current world plugin (Section \ref{subsec:current}).  When a bottom-track solution is unavailable due to sensor altitude, the sensor automatically switches to water tracking, if enabled. The water-track solution is computed in a manner similar to the bottom-track solution, except that beam-specific velocities are calculated 
relative to the global ocean current. As with bottom tracking, noise-free velocity is published by the sensor as a Gazebo message, and noisy velocity is published by the plugin as a \ac{ROS} message. The \ac{ROS} message metadata differentiates between water tracking and bottom tracking solutions.

Our ocean current profiling allows the \ac{DVL} to compute stratified relative velocities at intervals out to its maximum range. The \ac{DVL} subscribes to the Gazebo stratified current message and uses these values as the basis for sensed interval velocities.  For each beam, the sensor computes a real-world depth for the center of each range interval. Interpolated currents are computed from the stratified current at each interval depth and used to derive noise-free scalar beam velocities.

A current-profiling \acs{DVL} can usually provide either a combined velocity solution or a set of four beam-specific velocities for each interval. The DAVE \ac{DVL} plugin uses a custom \ac{ADCP} \ac{ROS} message for this purpose. If the profile is to be presented as a series of combined velocities, they are computed in the sensor frame by adding Gaussian noise to the beam-specific interval velocities and computing sensor interval velocities in the same manner as the bottom-track and water-track solutions.  If, on the other hand, the profile is to be presented as a series of beam-specific velocity 4-tuples, each value is computed by adding Gaussian noise to the scalar interval velocity and using the beam's unit vector (i.e., its orientation) to compute a sensor-frame velocity. The plugin publishes the computed profile and its parameters (e.g., beam orientations, interval sizes, number of intervals) using the \ac{ADCP} message.

\subsection{Underwater lidar}\label{subsec:laser}


Underwater lidar sensors provide \ac{UUV} platforms with the ability to obtain high-precision 3D imaging of objects at relatively close range. Sensors such as the 3d-at-Depth Underwater Laser system are commercially available and used in a variety of applications from oil and gas drilling to environmental surveys~\cite{filisetti2018developments}. However, although Gazebo-based models for terrestrial lidar systems are available, no comparable model for underwater lidar currently exists to our knowledge.

\begin{figure}[thbp]
    \centering
    \includegraphics[width=0.45\textwidth]{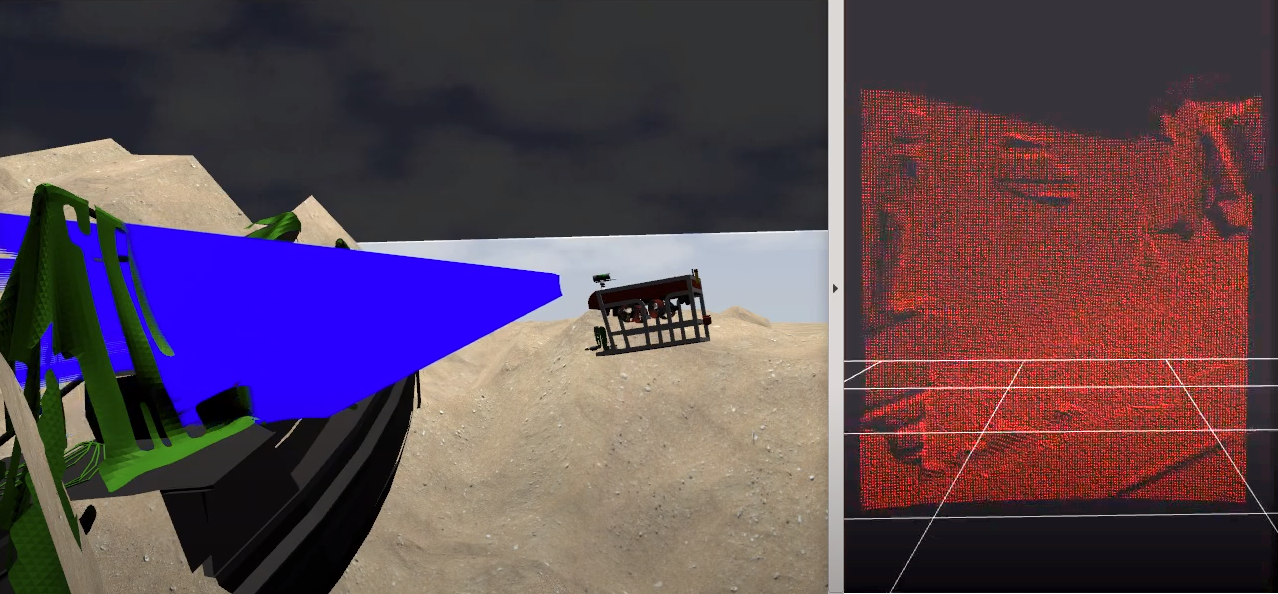}
    \caption{Left: underwater lidar scanning a shipwreck. Right: visualization of the resulting point cloud.}
    \label{f:uwlidarwreck}
\end{figure}

To address this deficiency, we adapt Gazebo's GPURaySensor to provide a first-order approximation of an underwater 3D pulse lidar sensor, modeled on the attributes of commercial off-the-shelf underwater lidar devices. This sensor (Fig.~\ref{f:uwlidarwreck}) produces a fixed field of $145 \times 145$ rays covering a $30 \times 30$ sector, and is set to a default range of 20 meters. The resolution can be adjusted according to the capabilities of the system hardware. The default resolution produces 10 simulated points per ray, resulting in a point cloud of $1450 \times 1450$ points. The default range is 20 meters, corresponding to relatively clear water conditions, but this can also be adjusted as needed according to the turbidity and light conditions. 

Because the sensor produces a fixed sector of points (unlike common lidar systems for land-based applications, which produce a point cloud that covers a $360^{\circ}$ sweep using a set of spinning lasers), the sensor is mounted within a heavy cylinder that responds to pan and tilt commands. This mount is capable of a $350^{\circ}$ panning motion and $60^{\circ}$ tilt, which extends the total reachable view to $360^{\circ} \times 90^{\circ}$.

\subsection{Multibeam forward-looking sonar (FLS)}\label{subsec:fls}


A novel physics-based simulated sonar, which takes advantage of \ac{GPU} parallelization, is developed for a multibeam forward-looking echo sounder~\cite{choi2021multibeam}. Unlike previous simulated sensors, which projected point clouds of the scene onto a 2D image to add per-pixel shading by image processing, our new simulated sensor calculates the returned intensities of each beam and the interference among multiple beams. As a result, the acoustic characteristics capture time and angle ambiguities and speckle noise, more realistically resembling an actual sensor. Additionally, the raw sonar data (intensity-range data, the A-plot) is produced and published as a ROS message.

\begin{figure}[thbp]
    \centering
    \includegraphics[trim={100px 0 50px 0},clip,width=0.2\textwidth]{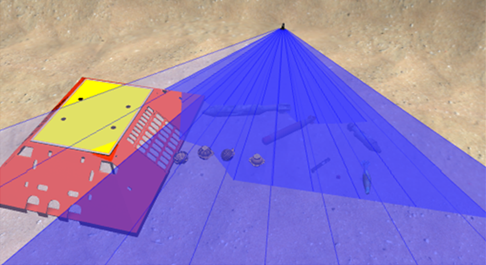}
    \includegraphics[width=0.28\textwidth]{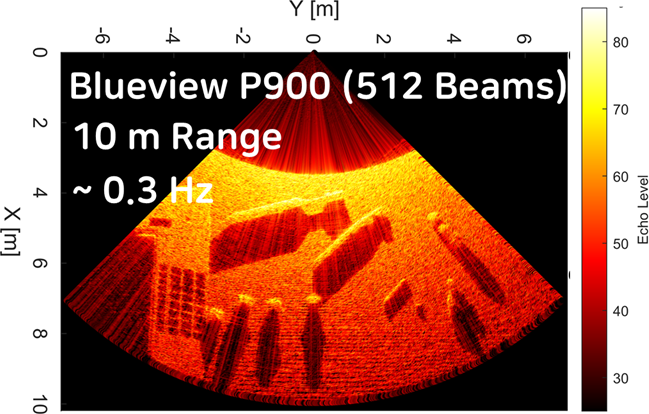}
    \caption{Local area search scenario demonstration using the simulated multibeam sonar.}
    \label{f:multibeam}
\end{figure}

The point scattering model is adopted and calculated for each ray within the beams, in parallel using \ac{GPU} cores. The spectrum of each ray is formulated in the frequency domain, including the target distance, reflectivity with incident angles, source level, and beam pattern for ray scatterers. A refresh rate of up to 10 Hertz is achieved for a sensor with 512 beams, on a workstation with an Intel i9-9900K 3.6 GHz processor and an Nvidia GeForce RTX 2080Ti \ac{GPU} (Fig.~\ref{f:multibeam}).

\section{Scenarios}\label{sec:scenarios}


To provide a useful environment for the development of autonomous underwater manipulation, we focus on modeling aspects that illustrate unique challenges in the deep ocean.
First, example scenarios are provided for common underwater manipulation use cases. Motion planning tools are integrated to offer an easier manipulation interface.
Second, the scenarios begin to address several challenges, including coping with instability in the presence of currents, utilizing acoustic sensors and bathymetry for terrain-aided navigation, piloting visual occlusion of buried objects and the effect of viscosity on contact physics, and operating in large-scale worlds.

\subsection{Electrical flying lead}\label{subsec:flying_lead}

\begin{figure*}[thbp]
    \begin{center}
        \begin{subfigure}{.33\textwidth}
            \includegraphics[height=\demofigheight]{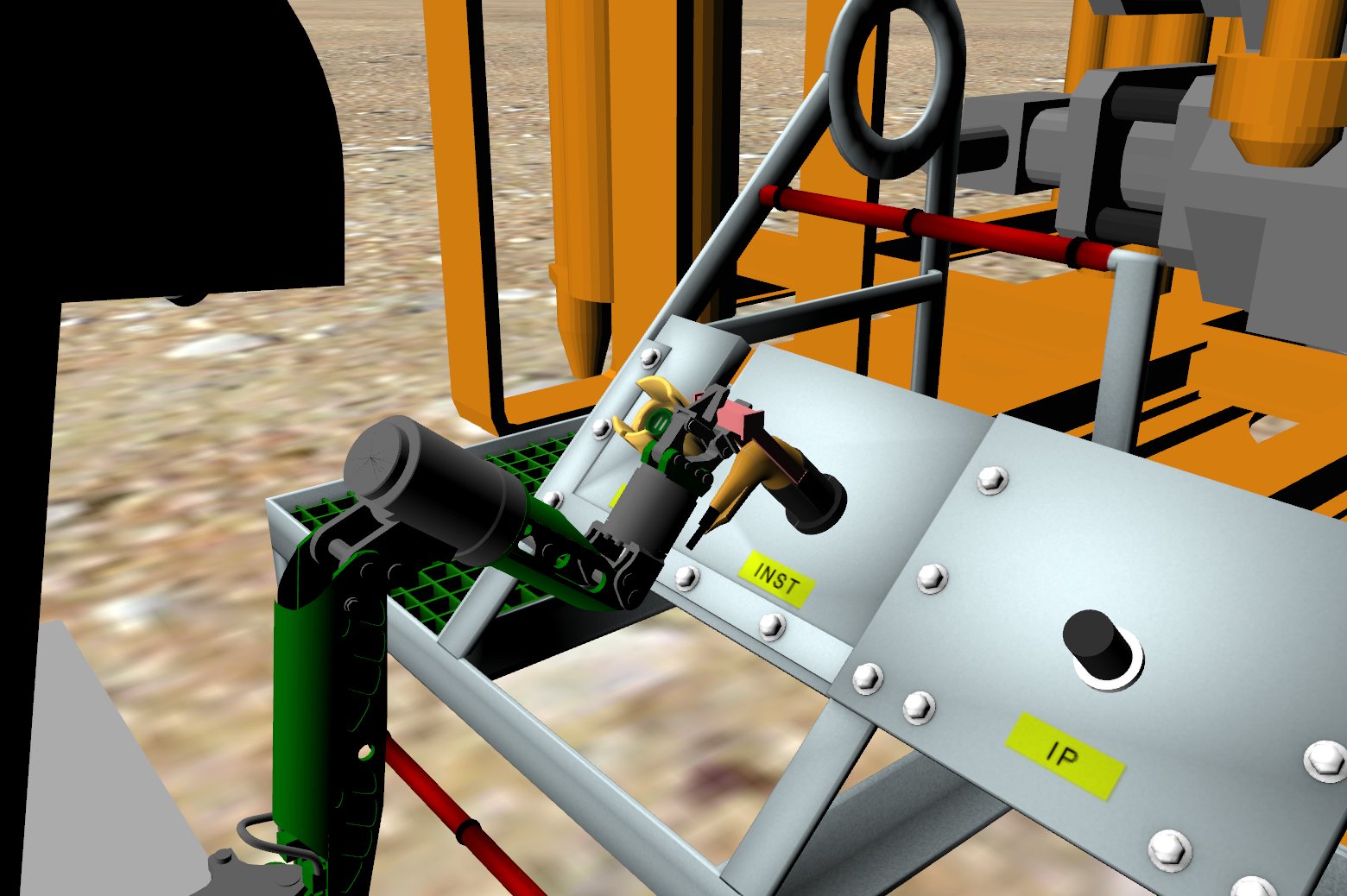}
            \caption{}
            \label{f:flying_lead}
        \end{subfigure} \hfill
        \begin{subfigure}{.30\textwidth}
            \includegraphics[trim={0 0 0 50px},clip,height=\demofigheight]{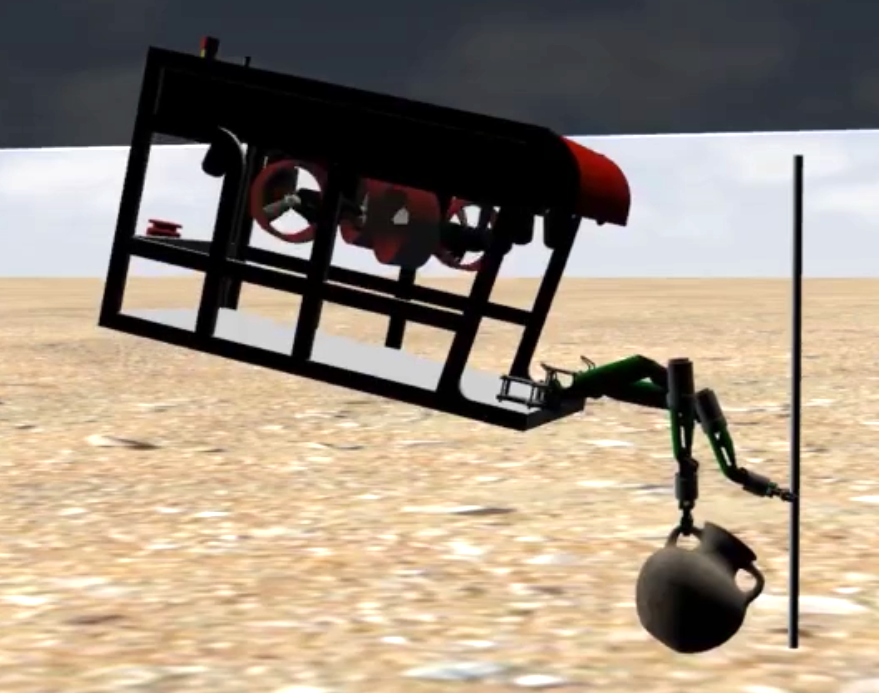}
            \caption{}
            \label{f:bimanual}
        \end{subfigure} \hfill
        \begin{subfigure}{.32\textwidth}
            \includegraphics[trim={100px 0 200px 0},clip,height=\demofigheight]{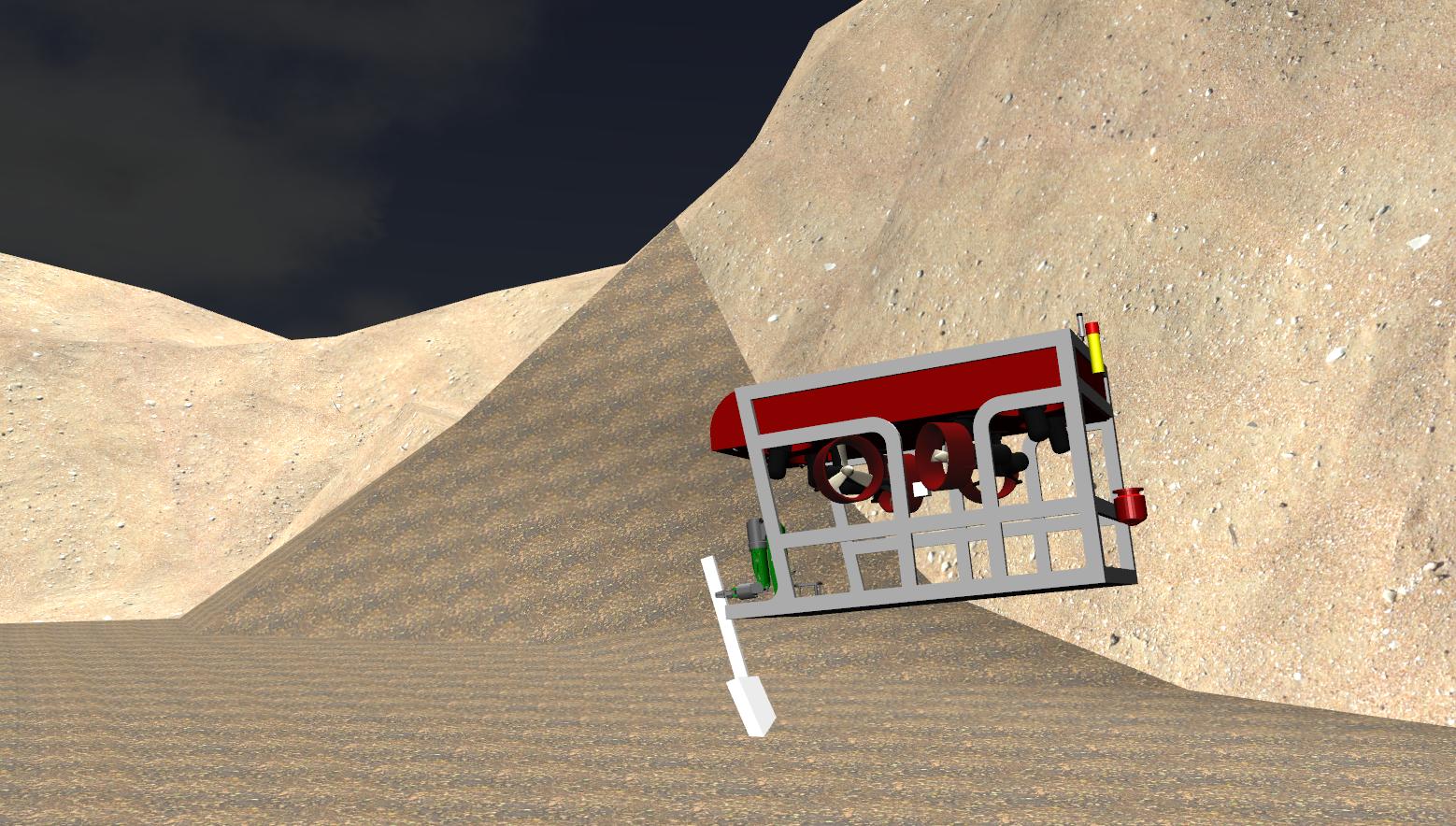}
            \caption{}
            \label{f:mud_demo}
        \end{subfigure} \hfill%
    \vspace{-4mm}
    \caption{(a) Flying lead join. (b) Simple bimanual manipulation example. The RexROV stabilizes itself on a fixed pole while grasping a vase. (c) Manipulating object with damping and visual occlusion.}
    \vspace{-5mm}
    \end{center}
\end{figure*}

A number of potential operational scenarios may require the robot to manipulate connectors of various forms.  Examples include maintenance or repair of underwater machinery and working with underwater cables. In many cases, these missions require the robot to connect or disconnect plugs to or from receptacles (e.g., an electrical lead to be plugged into a socket). To facilitate realistic simulation of these scenarios, we developed an electrical flying lead model plugin that utilizes a temporary joint connecting a plug model and a compatible receptacle model. Fig.~\ref{f:flying_lead} depicts a RexROV connecting a female plug to a male receptacle using the plugin. 

The plug-and-socket mechanism was developed as a model plugin associated with a receptacle model. Parameters such as alignment tolerances and required insertion and extraction forces are specified in the plugin's \ac{SDF}.  The \ac{SDF} is also used to specify the model and link names of the plug to be connected to the receptacle.  This allows the world to include multiple receptacles compatible with a single plug. A receptacle can only be associated with a single plug at present.

\begin{figure}[htbp]
  \centering
  \begin{tikzpicture}[node distance=1.5cm]
    \node (start) [node] {Start};
    \node (free) [node, below of=start] {Free};
    \node (fixed) [node, below of=free, xshift=-1.1cm] {Fixed};
    \node (joined) [node, below of=free, xshift=1.1cm] {Joined};
    \draw [arrow] (start) -- (free);
    \draw [arrow, text width=3cm] (free) [out=-15, in=90] to 
      node[anchor=south west]{Aligned for 2 seconds \& not recently freed} 
      (joined);
    \draw [arrow] (joined) [out=230, in=310] to
      node[anchor=north, yshift=-1mm]{Force toward socket}
      (fixed);
    \draw [arrow, text width=3cm, align=right, xshift=-3mm] (fixed) [out=90, in=200] to 
      node[anchor=south east]{Force away from socket}
      (free);
  \end{tikzpicture}
  \caption{Plug-and-socket state machine for the flying lead.}
  \label{f:state_machine}
\end{figure}
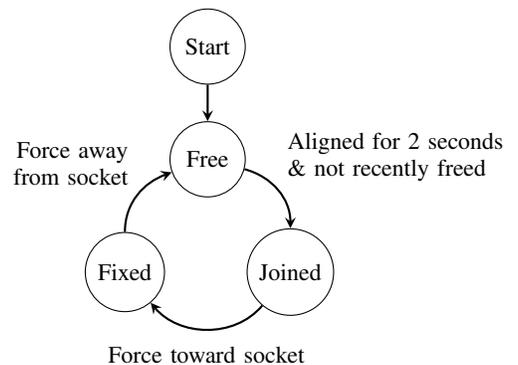

The plugin is implemented as a state machine that transitions between free, joined, and fixed states (Fig. \ref{f:state_machine}). In the free state, the plug and receptacle are not connected and can move freely. When the plug and receptacle are in close proximity and their link frames aligned (i.e., within the specified angular tolerances), the plugin transitions to the joined state and dynamically creates a temporary prismatic joint to bind the plug to the socket. This 1DOF joint allows the plug to move towards or away from the receptacle. 

When the plug is inserted such that the receptacle exerts sufficient force along the plug frame's $X$ axis (i.e., exceeding the specified insertion force), the plugin transitions to the fixed state, and the prismatic joint limits are set to zero to lock the plug in position. To release the plug, force must be applied to the plug, e.g., by a gripper. Sufficient force along the said $X$ axis (i.e., exceeding the extraction force) will ``release'' the plug by disconnecting and deleting the prismatic joint.

In the joined and fixed states, applied linear force is published as a \ac{ROS} Vector3Stamped message.  In the joined state, force applied to the plug by the receptacle is published. In the fixed state, force applied by the robot's gripper is published. Though the force is published as a 3D vector in the plug's frame, only the $X$ component is used to trigger state transitions. Because of the sources of the applied force, both insertion and extraction forces are positive in the $X$ direction.

\subsection{Bimanual manipulation}\label{subsec:biman_manip}


We can extend the single-arm manipulation in Section~\ref{subsec:flying_lead} to scenarios such as sample collection, debris cleanup, and vehicle stabilization. While combining these tasks can be challenging with a single arm, the robot's capabilities can be significantly expanded with two arms.

To that end, we added a second arm and configured both with the MoveIt framework~\cite{coleman2014}. MoveIt is an open-source manipulation platform that integrates multiple motion planning libraries and is built on ROS. It provides the means to add kinematics options, motion planning, and collision checking.
For each arm, we configured the robot using the MoveIt assistant and created gripper and arm controllers, as an arm controller might not be appropriate for a gripper. We implemented two controller types for the two arms. For simple movements, a position controller was defined for each gripper and arm. For effort and force feedback-dependent scenarios, an effort controller was built for each gripper and arm.

We provide examples using effort controllers, as they allow for more robust grasping than position controllers, yet produce reasonable arm movements in stabilization scenarios. PID tuning was performed with the arm under hydrodynamic and hydrostatic forces. These values can be revised for other simulations. While there are different ways to simulate grasping, we found that adjusting friction and effort values for the gripper allowed us to consistently grasp objects.

Examples in DAVE include a single arm scenario, multiple separately addressable arms in a single environment, and two arms that can be commanded simultaneously, though with a higher risk of collision.
This allows the user to easily command the arms in several ways, plan paths with collision avoidance, and receive joint effort feedback. Fig.~\ref{f:bimanual} shows a robot's final pose in a provided scenario grasping two objects, one being static, demonstrating the potential for stabilization when using the ocean currents plugin (Section~\ref{subsec:current}).

\subsection{Terrain-aided navigation}\label{subsec:terrain_aided_nav}


\ac{TAN} is an underwater localization method that compares sensor data with a terrain map. A vehicle's position is based on bathymetric measurements. Taking advantage of our simulated sensors, we outline a scenario to equip a vehicle for \ac{TAN}. We split the \ac{TAN} method into three parts: feature-based navigation, dead reckoning, and positional error reset.

Feature-based navigation is our focus and takes advantage of two simulated sensors. The multibeam sonar (Section~\ref{subsec:fls}) can be used particularly to identify high frequency seafloor features or to generate a point cloud for landmarks. We also created a scenario for using a \ac{DVL} (Section~\ref{subsec:dvl}) to estimate a seabed gradient map. These two sensors combine high and low frequency sensing, which can be used to identify a range of seafloor features for navigation.

In the absence of detailed terrain maps or while navigating in  conditions unfavorable to terrain following, dead reckoning can be used with sensors in DAVE, in particular the \ac{INS}, the \ac{DVL}, and the \ac{USBL}.
The \ac{INS} allows for inertial and pressure measurements, with the goal of minimizing the error between expected and reported sensor values while following a planned path.
The \ac{DVL}'s bottom tracking is commonly used in \ac{TAN} to minimize deviations from a set heading. These values can be fused with \ac{INS} values in GPS-denied or map-deprived scenarios.
In more controlled scenarios, the \ac{USBL} can be used to track the vehicle's relative position regardless of map or GPS availability.

For both feature-based navigation and dead reckoning, positional error reset typically relies on a GPS. Using the GPS plugin in DAVE allows for two options. The first is system validation, using the GPS to measure positional path errors. The second is position re-calibration. GPS is usually unavailable in \ac{TAN}-requiring scenarios but can be used during a brief resurfacing to determine the point-to-point positional error. A ground truth can be determined via the Gazebo {\codefont|get_model_state|} service.


\subsection{Surface occlusion and effects}\label{subsec:occlusion}


We provide scenarios for modeling the interaction between vehicles and manipulation targets, with different seafloor makeups and varying degrees of mechanical (Section~\ref{subsec:distortion}) and biological fouling.
Our methods begin to address: changes in shape, color, and texture due to marine growth; visual occlusion due to permeable bottom composition; and changes in reaction forces due to partial burial in a viscous medium.

We exploit the structure of the model-defining \ac{SDF} to select the sensor types to target with a particular effect.  Visual and collision elements are defined separately, and physical properties can be adjusted on a model-by-model basis.  Some sensors and plugins operate strictly on the visual elements: those relying on the rendering pipeline, notably the GPURaySensor, CameraSensor, and DepthCameraSensor.  Others, including Gazebo's RaySensor and SonarSensor, interact with the collision elements using the physics engine.  Physical models, such as the vehicle’s chassis and manipulator arm, interact with the collision elements under most circumstances.

We use a simple offset between the visual and collision \ac{SDF} elements of the seafloor model to create a visual occlusion effect, similar to objects being partially submerged in a silty surface.  This method has the benefit of coming at no computational cost, and is effective in a challenging manipulation scenario.
A similar method is employed to ``grow'' biological fouling agents onto a submerged object.
The visual element of the model is scaled up by a small percentage in all axes, and a suitable color or texture mapping is applied.
For scaling complex objects or creating multiple effects across a section of seafloor, which may require more flexibility than in the \ac{SDF}, we modify meshes using 3D modeling software~\cite{blender}.

In addition to visual occlusion, we simulate damping effects on object motion caused by partial submersion in a viscous medium.  We create a new model from the seafloor mesh to represent an area of mud (or other bottom types).  This model must contain a collision element for interacting with other objects, but must also be able to intersect
those objects. This is accomplished by setting {\codefont|collide_without_contact|} to {\codefont|true|} in the collision \ac{SDF}. A visual element is optional.  Fig.~\ref{f:mud_demo} shows the medium resisting a dragging motion.

The motion damping is implemented via the Gazebo MudPlugin~\cite{koenig2004gazebo}, which creates a temporary joint between the mud model and all links tagged to interact with it, and a contact sensor.
Because this implementation uses the \ac{SDF} collision element, unlike the simpler visual method above, sensors that use collision checking will react to the new surface. If this is not desired, it is possible to filter collisions using a bitmask.

\subsection{Integrated scenario}\label{subsec:integrated_scenario}

\begin{figure}[htbp]
    \centering
    \includegraphics[width=0.48\textwidth]{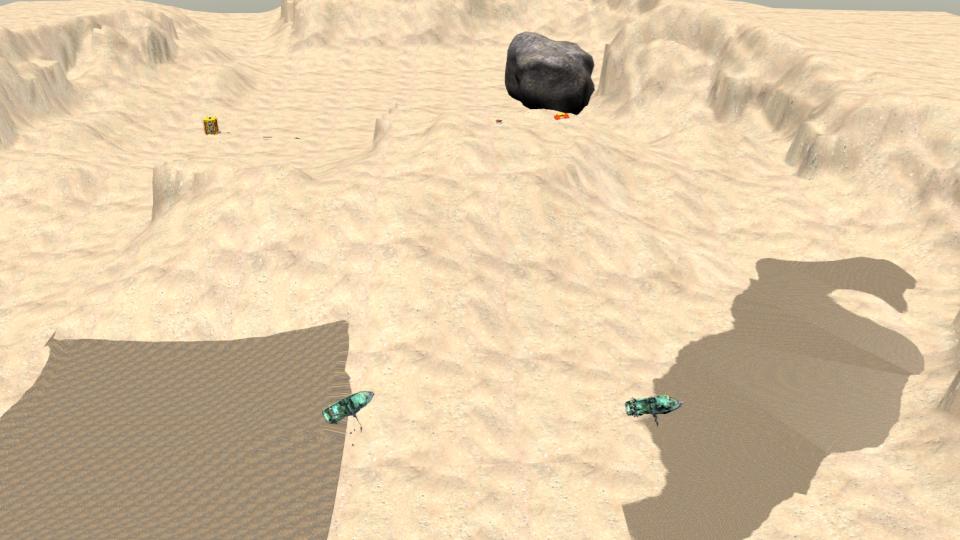}
    \caption{Integrated scenario with electrical flying lead and bimanual manipulation (top left), surface occlusion (shaded areas at the bottom), and distorted objects (scattered).}
    \label{f:integrated}
\end{figure}

We present a combined scenario that integrates multiple aforementioned examples into a single world, to demonstrate a more complex configuration in which multiple tasks can be carried out.
This is an example of many permutations that could be created with DAVE, allowing for the training and evaluation of different missions.

Fig.~\ref{f:integrated} shows a downscaled bathymetric map of the Santorini caldera, featuring various elevations. The seafloor is augmented by two mud pits, one with variable depths and the other of a constant depth that follows the terrain. A number of objects, distorted and undistorted, are placed in the scene for the robot to interact with.
A mission is configured to couple the electrical flying lead (Section~\ref{subsec:flying_lead}) and the bimanual manipulation (Section~\ref{subsec:biman_manip}), such that a dual-arm RexROV is used to manipulate a plug into a receptacle (Fig.~\ref{f:cover}).

The bathymetric map is relatively large even when downscaled. This resembles actual scenarios but also means time-consuming navigation. To facilitate quicker evaluations, we provide a script to teleport multiple vehicles between stations.

\section{Conclusion and future work}\label{sec:conclusion}

We introduced environments and sensors that are the common building blocks of many underwater applications and illustrated their use in various scenarios.
However, the community is still far from arriving at a general-purpose simulator.

On the environment front, other types of geospatial data, such as chemical and biological composition, may be of interest to some applications.
On the sensor front, the multibeam forward-looking sonar may be adapted to other types, such as a side scan sonar, which is more affordable and widely available.
To take advantage of recent developments in visual perception on land, simulating underwater cameras can be a significant step to transfer visual learning algorithms to underwater.

\section*{Acknowledgment}

\small{
DAVE would have been impossible without the basis of existing open source software, especially the UUV Simulator \cite{manhaes2016uuv} and the {\codefont|ds_sim|} from the WHOI Deep Submergence Lab \cite{vaughn_dssim}.
In addition, we thank
Derek Olson and Andy Racson at Naval Postgraduate School (NPS) for theoretical contributions and mathematical implementations;
Michael Jakuba at WHOI for the original dynamic bathymetry implementation;
Youssef Khaky for software contributions;
Ryan Lee at NPS and Cole Biesemeyer and Althone Torregosa at Open Robotics for 3D modeling.

This research was developed with funding from the Defense Advanced Research Projects Agency (DARPA).

Reference herein to any specific commercial product, process, or service by trade name, trademark or other trade name, manufacturer or otherwise, does not necessarily constitute or imply endorsement by DARPA, the Defense Department or the U.S. government, and shall not be used for advertising or product endorsement purposes.

}

\bibliographystyle{IEEEtran}
\bibliography{paper, bbing_master}

\end{document}